\documentclass{article}
\PassOptionsToPackage{numbers, compress}{natbib}
\usepackage[final]{nips_2017}
\usepackage[english]{babel}
\usepackage[T1]{fontenc}

\usepackage{amsmath}
\usepackage{graphicx}
\usepackage[colorinlistoftodos]{todonotes}
\usepackage[colorlinks=true, allcolors=blue]{hyperref}
\usepackage{natbib}
\usepackage{kky}

\usepackage{hyperref}       
\usepackage{url}            
\usepackage{booktabs}       
\usepackage{amsfonts}       
\usepackage{nicefrac}       
\usepackage{microtype}      
\usepackage{multirow}
\usepackage{algorithm}
\usepackage{algorithmic}
\usepackage{mathrsfs}
\usepackage{graphicx}
\title{Few-shot Learning with Meta Metric Learners}
\author{
  Yu Cheng$^1$ \hspace{0.7em} Mo Yu$^2$ \hspace{0.7em} Xiaoxiao Guo$^2$ \hspace{0.7em} Bowen Zhou$^3$  \\
  $^1$ Microsoft AI \& Research,\hspace{1em} $^2$IBM Research AI,\hspace{1em} $^3$JD AI Research \\
  \texttt{yu.cheng@microsoft.com}\hspace{1em} \texttt{\{yum,xiaoxiao.guo\}@ibm.com} \hspace{1em}  bowen.zhou@jd.com
}

\def \scs {SCS}

\begin{document}
%
%
%
\maketitle

\begin{abstract}
Few-shot Learning aims to learn classifiers for new classes with only a few training examples per class. Existing meta-learning or metric-learning based few-shot learning approaches are limited in handling diverse domains with various number of labels. The meta-learning approaches train a meta learner to predict weights of homogeneous-structured task-specific networks, requiring a uniform number of classes across tasks.
The metric-learning approaches learn one task-invariant metric for all the tasks, and they fail if the tasks diverge.
We propose to deal with these limitations with meta metric learning. Our meta metric learning approach consists of task-specific learners, that exploit metric learning to handle flexible labels, and a meta learner, that discovers good parameters and gradient decent to specify the metrics in task-specific learners. Thus the proposed model is able to handle unbalanced classes as well as to generate task-specific metrics. We test our approach in the `$k$-shot $N$-way' few-shot learning setting used in previous work and new realistic few-shot setting with diverse multi-domain tasks and flexible label numbers. Experiments show that our approach attains superior performances in both settings.
\end{abstract}

\section{Introduction}

Supervised deep learning methods have been successfully applied to many applications such as computer vision, speech recognition and natural language processing. In practice, those methods usually require large amount of labeled data for model training, in order to make the learned model generalize well.
However, collecting sufficient amount of training data for each task  needs a lot of human-labeling work and the process is time-consuming.

Few-shot learning~\cite{li2006one} was proposed to learn classifiers for new classes with only a few training examples per class.
Two key ideas of few-shot learning are data aggregation and knowledge sharing.
First, though each few-shot learning task may lack sufficient training data, the union of all the tasks will provide significant amount of labeled data for model training. Therefore the model training and prediction on a new coming few-shot could benefit from all the learned tasks. Secondly, the experiences of learning model parameters for a large number of tasks in the past will assist the learning process of the incoming new task.
The few-shot learning idea has recently been combined with the deep learning techniques in two main lines of works: (1) learning metric/similarity from multiple few-shot learning tasks with deep networks \citep{koch2015siamese,vinyals2016matching}; and (2) learning a meta-model on multiple few-shot learning tasks, which could be then used to predict model weights given a new few-shot learning task \cite{kaiser2017learning,ravi2017optimization,munkhdalai2017meta}.

The aforementioned deep few-shot learning models usually are applied to the so called ``$k$-shot, $N$-way'' scenario, in which each few-shot learning task has the same $N$ number of class labels and each label has $k$ training instances.
%
However, such ``$N$-way'' simplification is not realistic in real-world few-shot learning applications, because different tasks usually do not have the same number of labels.
Existing meta-learning approaches build on the ``$N$-way'' simplification to let the meta-learner predict weights of homogeneous-structured task-specific networks. If we allow different tasks work with different number of labels, the task-specific networks will be  heterogeneous.  Heterogeneous-structured task-specific networks complicate the weight prediction of the meta-learner. To the best of our knowledge, none of the existing meta-learning based few-shot learning approaches could resolve this issue.
Although the metric-learning approaches could alleviate the variations on class labels, they suffer from the limitation of model expressiveness: these methods usually learn a task-invariant metric for all the few-shot learning tasks. However, because of the variety of tasks, the optimal metric will also vary across tasks. The learned task-invariant metric would fail if the tasks diverge. 

Moreover, in real world applications, the few-shot learning tasks usually come from different domains or different resources. For example, for sentiment classification of product reviews, we could have products from different product departments on an e-commerce platform like Amazon. For an machine learning cloud service, there may be different clients submitting training data for their own business tasks. The tasks from different clients may deal with different problems, such as spam detection or sentiment classification.
In such few-shot learning scenario, the two aforementioned issues of existing few-shot learning approaches will become more serious: the numbers and meanings of class labels may vary a lot among different tasks, so it will be hard for a meta-learner to learn how to predict weights for heterogeneous neural networks given the few-shot labeled data; and different tasks are not guaranteed to be even closely related to each other, so there will unlikely exits a uniform metric suitable for all the tasks from different domains or resources.

\begin{figure}
\includegraphics{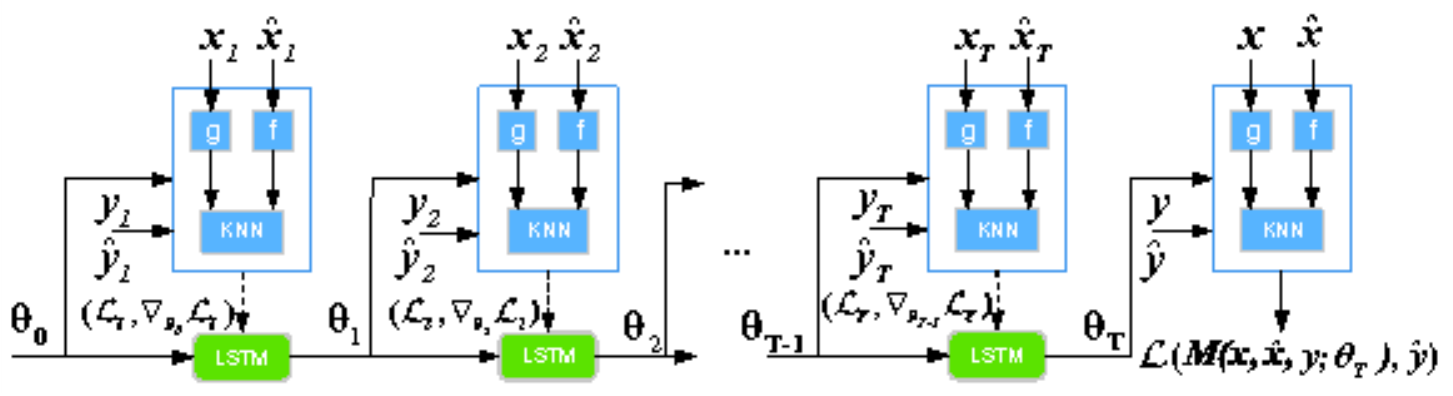}
\caption{Computational graph of the forward pass of meta metric learner. Each ($x_{i}, y_{i})$ is the $i^{th}$ batch sampled from $D_{train}$ and ($x$, $y$) are all the samples of $D_{train}$. Analogously, each ($\hat{x}_{i}, \hat{y}_{i})$ is the $i^{th}$ batch sampled from $D_{test}$ and ($\hat{x}$, $\hat{y}$) are all the samples of $D_{test}$. The dashed arrows indicate that the gradient is not back-propagated though that step when training the meta metric learner. We refer to the matching network learner as $M$ with two encoders named $g$ and $f$, and $M(x,\hat{x},y;\theta_{T})$ is the output of learner $M$. }
\label{fig:arc}
\end{figure}

We propose to deal with the limitations of previous work with meta metric learning. The model consists of two main learning modules (Figure \ref{fig:arc}). The meta learner that operates across tasks uses an LSTM-based architecture to discover good parameters and gradient decent in task-specific base learners. The base learners exploit Matching Networks~\cite{vinyals2016matching} and parameterize the task metrics using the weight prediction from the meta-learner. Thus the proposed model is now able to handle unbalanced classes in meta-train and meta-test sets as the usage of Matching Network as well as to generate task-specific metrics leveraging the weight prediction of the meta-learner given task instances.

We make the following contributions:
(1) we improve the existing few-shot learning work to handle various class labels; (2) we further enable the model to learn task specific metrics via training a meta learner
and (3)  we are the first to investigate few-shot deep learning methods in the text domains.
The contributions (1) and (2) make our approach suit better to the real few-shot learning scenarios where different tasks have various numbers of class labels and could come from different domains.
We test our approach in the classic ``$k$-shot $N$-way''  few-shot learning setting following previous work
and a new but more realistic few-shot setting with diverse multi-domain tasks and flexible label numbers.
Experiments show that our approach attains superior performance on both settings.

\section{Backgrounds on Few-Shot Learning}
\label{sec:backgrounds}

Few-Shot Learning (FSL)~\citep{li2006one,miller2000learning} aims to learn classifiers for new classes with only a few training examples per class. Bayesian Program Induction~\cite{lake2015human} represents concepts as simple programs that best explain observed examples under a Bayesian criterion.  Siamese neural networks rank similarity between inputs~\cite{koch2015siamese}. Matching Networks~\cite{vinyals2016matching} map a small labeled support set and an unlabeled example to its
label, obviating the need for fine-tuning to adapt to new class types. Siamese neural networks and Machine Networks essentially learn one metric for all the tasks, but to use only one metric is sub-optimal when the tasks are diverse. Recently, an LSTM-based meta-learner~\cite{ravi2017optimization} learns the exact optimization algorithm used to train another learner neural network classifier for the few-shot learning problems. However, it requires a uniform number of classes across tasks.
Our FSL approach could handle the challenges of diverse tasks with flexible labels.

\subsection{Matching Network}

Matching network~\cite{vinyals2016matching} is a recent model developed for few-shot learning in computer vision applications. We adapt it to handle flexible class numbers in our Meta Metric Learner for both single-task and multi-task few-shot learning problems.
Matching networks consist of two neural network (with shared parameters) as embedding functions and an augmented memory. The embedding functions, $f()$ and $g()$, map a review paragraph $x \in X$ to a $N$-length vector, i.e., $f, g: X \rightarrow \mathbb{R}^{N}$.  The augmented memory stores a support set $S=\{(x_{i},y_{i})\}^{k}_{i=1}$, where $x_{i}$ is the supporting instance and $y_{i}$ is its corresponding label in a one-hot format. The matching networks explicitly define a classifier $c_{S}$ conditioned on the supporting set. For any new data $\hat{x}$, the matching network predicts its label via a similarity function $\alpha(.,.)$ between the instance and  the support set:
\begin{equation}
y = P(.|\hat{x}, S) = \sum_{i=1}^{k} \alpha(\hat{x}, x_{i};\theta) y_{i}.
\end{equation}
Specifically, we define the similarity function to be a softmax distribution given the inter-product between the test instance $\hat{x}$ and the supporting instance $x_{i}$, {\emph i.e.}, $\alpha(\hat{x}, x_{i};\theta) = \exp(f(\hat{x}) \cdot g(x_{i})) / \sum_{j=1}^{k} \exp(f(\hat{x}) \cdot g(x_{j}))$, where $\theta$ are the parameters of the embedding functions $f$ and $g$. Thus, $y$ is a distribution over the supporting set's labels $\{y_{i}\}_{i=1}^{k}$.
We choose $f$ to be convolutional neural networks following ~\cite{kim:2014:EMNLP2014,johnson2016supervised}.

\paragraph{Training Objective}
The original training objective of Matching network is specialized to match the test condition to the training condition for few-shot learning settings.
We first sample a few-shot dataset $D$ from all available datasets $T$, $D\sim T$. For each class in the sampled dataset $D$, we sample $k$ random instances in that class to construct a support set $S$, and sample a batch of training instances $B$ as training examples, i.e., $B,S\sim D$.
The objective function to optimize the embedding parameters is to minimize the prediction error of the training samples given the supporting set as the follow:
\begin{equation}
\mathop{\mathbb{E}}_{D\sim T} \Big[ \mathop{\mathbb{E}}_{B,S\sim D} \big[ \sum_{(x,y)\in B} \log(P(y|x,S;\theta))\big] \Big].
\end{equation}
The parameters of the embedding function, $\theta$, are optimized via stochastic gradient descent methods.

\subsection{Meta Learning}
Meta-learning (also known as Learning to learn) has a long history~\cite{thrun1998learning, schmidhuber1997shifting}.
The key idea is framing the learning problem at two levels: the first is the quick acquisition of knowledge from each separate task presented and the second is accumulating these knowledge to learn the similarities and difference across all tasks.
A recent approach to meta-learning \cite{andrychowicz2016learning} casts the hand-designed optimization algorithm as a learning problem, and trains an LSTM-based meta-learner to predict model parameters. An LSTM-based meta-learner~\cite{ravi2017optimization} was applied to few-shot learning tasks.

The procedure of training a learner with parameters $\theta$ can be expressed as the problem of optimizing some loss function $\mathcal{L}(\theta)$ over some domain. The standard optimization algorithms are some variant of gradient descend:
\begin{equation}
\theta_{t+1} = \theta_{t} - \alpha_{t+1}\nabla \mathcal{L}(\theta_{t})
\end{equation}
where $\theta_{t}$ are the learner parameters after $t$ update steps, $\alpha_{t+1}$ is the learning rate at time step $t+1$, $\nabla \mathcal{L}(\theta_{t})$ is the gradient of the loss function with respect to parameters $\theta_{t}$, and $\theta_{t+1}$ are the updated parameters of the learner. The LSTM-based meta-learner leverages that the update of learner parameters resembles the update of the cell state in an LSTM~\cite{hochreiter1997long}:
\begin{equation}
c_{t+1} = f_{t+1}\odot c_{t} + i_{t+1}\odot \widetilde{c}_{t+1}
\end{equation}

If we set the cell state of LSTM to be the parameters of the learner, i.e. $c_{t}=\theta_{t}$, the candidate cell state $\widetilde{c}_{t+1}=-\nabla \mathcal{L}(\theta_{t})$, the output of forget gate $f_{t+1} = 1$, and the output of input gate $i_{t+1} = \alpha_{t+1}$, these two update procedures are completely the same.
$i_{t+1}$ and $f_{t+1}$ determine how the meta-learner updates the parameters of learner. Thus, an LSTM can be trained as a meta-learner to learn an update rule for training a learner (such as a neural network).


\section{Meta Metric Learner for Few-Shot Learning}
In this section, we first describe the meta metric-learner in a single-task setting. After that, we show it is easy to generalize the model in a multi-task learning setting, which relates to retrieve auxiliary sets from other sources/tasks.

\subsection{Meta Metric Learner}
Let's consider the meta-learning in few-shot setting, on a data resource $R$ with meta set $\mathcal{D}$, where $\mathcal{D}$ consists of three parts: $\mathscr{D}_{meta-train}$, $\mathscr{D}_{meta-validate}$ and $\mathscr{D}_{meta-test}$. Generally, in real applications, the number of classes in $\mathscr{D}_{meta-train}$ is different from that in $\mathscr{D}_{meta-test}$. Although the CNN base learner used in \cite{ravi2017optimization} is powerful to model image and text, it lacks an ability to handle unbalanced classes in train and test datasets in a straightforward way. Matching networks, on the other hand, as a trainable $k$NN and non-parametric algorithm by nature, can generalize easily to any new datasets, even if the number of classes are different. Hence, we apply the LSTM Meta-learner in \cite{ravi2017optimization} for few-shot learning tasks, but replace the CNN with Matching Networks as the base learner, so that it can tackle class-unbalanced few-shot learning problems. Matching networks is a kind of metric learning algorithm, and we train it using an LSTM-based meta-learner, so we call our method Meta Metric Learner.

Suppose we have a meta metric learner with parameters $\theta$ from the base learner $M$ and $\Theta$ from meta learner $R$. We can use the meta-train set $\mathscr{D}_{meta-train}$ to train the LSTM-based meta-learner. When using the trained meta-learner and meta-test set $\mathscr{D}_{meta-test}$ to update the parameters of base learner, i.e. a matching networks, it takes as input the loss and its gradient w.r.t parameters of matching networks. Thus we need labels of both train data and test data to compute them, which means we need more than one labeled data from meta-test set. This is fine in few-shot learning problems (using a subset of $\mathscr{D}_{meta-test}$), but violates the assumption of one-shot learning.
For class-unbalanced one-shot learning problems,  we propose to exploit resource from other tasks to update the parameters of matching networks. The auxiliary set is named $\mathscr{D}_{aux}$. How to choose appropriate auxiliary dataset is described in the next section.

The meta metric learning training algorithm is shown in Algorithm 1 and the framework is also shown in Figure 1. It is notable that even for few shot learning ($k \ne 1$) with multi-tasks resource, we do not necessarily to use the subset $\mathscr{D}_{sub}$. The auxiliary set $\mathscr{D}_{aux}$ can be used to support the training of base learner. In such situation, the learning algorithm could save resource as well as having a good generalization power.

\begin{algorithm}
\caption{Meta Metric Learner Training}
\label{alg1}
\begin{algorithmic}[1]
\REQUIRE ~~\\ Meta-train set $\mathscr{D}_{meta-train}$, Meta-test set $\mathscr{D}_{meta-test}$, auxiliary set $\mathscr{D}_{aux}$ or meta-test subset $\mathscr{D}_{sub}$,  Matching networks learner $M$ with parameters $\theta$, Meta-Learner $R$ with parameters $\Theta$

\STATE
\STATE****Meta-training****
\STATE $\Theta_{0}$ $\leftarrow$ random initialization
\FOR{$d=1,n$}
\STATE $D^{d}_{train}$,  $D^{d}_{test}$ $\leftarrow$ random sampled dataset from $\mathscr{D}_{meta-train}$
\STATE $\theta_{0} \leftarrow c_{0}$
\FOR{$t=1,T$}
\STATE $x_{t}$, $y_{t}$ $\leftarrow$ random batch sampled from $D^{d}_{train}$
\STATE $\hat{x}_{t}$, $\hat{y}_{t}$ $\leftarrow$ random batch sampled from $D^{d}_{test}$
\STATE $\mathcal{L}_{t} \leftarrow \mathcal{L}(M(x_{t}, \hat{x}_{t}, y_{t};\theta_{t-1}), \hat{y}_{t})$
\STATE $c_{t} \leftarrow R((\nabla_{\theta_{t-1}} \mathcal{L}_{t},\mathcal{L}_{t}));\Theta_{d-1})$
\STATE $\theta_{t} \leftarrow c_{t}$
\ENDFOR
\STATE $x$, $y$ $\leftarrow$ all samples from $D^{d}_{train}$
\STATE $\hat{x}$, $\hat{y}$ $\leftarrow$ all samples from  $D^{d}_{test}$
\STATE $\mathcal{L}_{test} \leftarrow \mathcal{L}(M(x, \hat{x}, y;\theta_{T}), \hat{y})$
\STATE Updating $\Theta_{d}$ using $\nabla_{\Theta_{d-1}}\mathcal{L}_{test}$
\ENDFOR
\STATE
\STATE****Updating Learner Parameters****

\STATE $D_{train}$,  $D_{test}$ $\leftarrow$ random sampled dataset from $\mathscr{D}_{aux}$ or $\mathscr{D}_{sub}$
\STATE $\theta_{0} \leftarrow c_{0}$
\FOR{$t=1,T$}
\STATE $x_{t}$, $y_{t}$ $\leftarrow$ random batch sampled from $D_{train}$
\STATE $\hat{x}_{t}$, $\hat{y}_{t}$ $\leftarrow$ random batch sampled from $D_{test}$
\STATE $\mathcal{L}_{t} \leftarrow \mathcal{L}(M(x_{t}, \hat{x}_{t}, y_{t}; \theta_{t-1}), \hat{y}_{t})$
\STATE $c_{t} \leftarrow R((\nabla_{\theta_{t-1}} \mathcal{L}_{t},\mathcal{L}_{t}));\Theta_{n})$
\STATE $\theta_{t} \leftarrow c_{t}$
\ENDFOR

\end{algorithmic}
\end{algorithm}

\subsection{Auxiliary Task Retrieval via Task-Level Matching Networks}
\label{ssec:method_retrieval}
In this section, we discuss how to choose the auxiliary set $\mathscr{D}_{aux}$. The intuition is, in many real world applications, we could get few-shot learning data from multiple tasks (sources), such as different text classification tasks on dialogues. Such resources from multiple tasks could significantly increase the training data for our few-shot learning method as well as the previous few-shot learning methods as described in Section \ref{sec:backgrounds}. However there is rarely a guarantee that those different tasks are related to each other.
When the tasks are from unrelated resources, it will be difficult for the existing few-shot learning methods to learn a good metric or a good meta-learner. In this case when adding more significantly unrelated training resources, the performance may decrease. To overcome this difficulty, given a target task for few-shot learning, we propose the following approach to select related tasks following \cite{N18-1109}.

Specifically, consider a list of $n$ tasks (such as a list of domains in sentiment analysis, or a list of alphabets in hand-writing recognition) $\mathcal{T} = \{T^1, T^2, ..., T^n\}$. From each task $T^i$ we could sample few-shot learning data sets $D^i_j$s. Because the $D^i_j$s are usually too small to reflect any statistical relatedness among them, our approach deal with the problem at the task-level with the following steps: (1) For each data resource $T^i$ we merge all the data sets $D^i_j$ together and train a matching network $M^i$ on it\footnote{On each $T^i$, this step works the same as the standard matching network approach under the single-task setting.}. (2) For the target task $T^{target}$, on its combined task we apply each model $M^i$ to get the accuracy $acc_{i\rightarrow k}$. Note that the accuracy scores are usually low but their relative magnitudes could reflect the relatedness between different tasks to $T^{target}$. (3) Finally we select the top $s$ tasks $T^i$ with the highest scores $acc_{i\rightarrow k}$ as the auxiliary set $\mathscr{D}_{aux}$ for the target task $T^{target}$.

\section{Experimental Results}
In this section, we execute experiment with $k$-shot learning in multi-task setting. The experiments are ran on three data sets: two text classification sets and one image classification set, comparing meta metric-learner model against several strong baselines. We first describe the datasets, experimental settings and baseline models.

\paragraph{Datasets} We introduce three datasets, which can be used in multi-task setting: 1) \textbf{Sentence Classification Service (SCS)}: The data set is from an on-line service which trains and serves text classification models for different clients for their business purposes. The number of total clients is 12, and the number of classes per clients ranges from 10 to 28.
2) \textbf{Omniglot}: the data comes with a standard split of 30 training alphabets with 964 classes and 20 evaluation alphabets with 659 classes. Each of these was hand drawn by 20 different people. The large number of classes (characters) with 20 data per class;  3) \textbf{Amazon Reviews}:  this corresponds a multi-domain sentiment classification data set from \cite{blitzer2007biographies}. The dataset consists of Amazon product reviews for 25 product types. Each review's label is a rating based on 1-5 stars\footnote{Data from \url{http://www.cs.jhu.edu/~mdredze/datasets/sentiment/}. The 3-star samples were removed due to their ambiguous nature \cite{blitzer2007biographies}.}.

\paragraph{Baseline Models} The first baseline we used is Matching Network. We implemented our own version of both the basic and the fully-conditional embedding (FCE) versions. The Second baseline is Meta-Learner LSTM. We implemented our own version according to \cite{ravi2017optimization}, which takes similar structures, dropping out and batch normalization.

\paragraph{CNN architectures} We exploit two different CNN architecture used in all the methods: 1) for text, we used a simple yet powerful CNN \cite{kim:2014:EMNLP2014} as the embedding function, which consists of a convolution layer and a max-pooling operation over the entire sentence for each feature map. The model uses multiple filters with varying window sizes $h (h=3,4,5)$. Word embedding are initialized with 100-dimensional Glove embeddings trained on 6B corpus from \citep{pennington2014glove}; 2) for image, the 2D CNN architecture in \cite{vinyals2016matching} is used, which consists of a stack of modules, a $3 \times 3$ convolution with 64 filters followed by batch normalization, a Relu non-linearity and $2 \times 2$ max-pooling.

\paragraph{Hyper-paramters} There are several hyper-parameters required for meta-learner, matching network and CNNs. All of them are tuned in the validation set.

\subsection{Multi-tasks/domains Setting}
\paragraph{Sentence Classification Service (SCS)}
On \scs, the data of each clients can be viewed as a task.
For each set, we randomly sample 50\% data into a meta-training set, 20\% into a meta-validation set, and assign the rest data into the meta-test set. For each sample, we did stop-word removal/tokenization with the CMU NLP tool~\cite{gimpel2011part} for the preprocessing. Since there are large number of clients on the service, random sampling auxiliary tasks has low chance to find related tasks. As a result, we use the method in Section \ref{ssec:method_retrieval} to retrieve top 10 related tasks from the 175-task pool. Then we train all the few-shot learning methods on the sampled tasks together with the meta-training set. The baseline, matching network is trained iteratively with one sampled task data each time. Obvious, meta-learner LSTM can not take the additional data and is only trained with the meta-training set.

After that, all the approaches are evaluated on the meta-test set with 5 samples per class. The validation set is used to adjust the hyper-parameter of the model. Results comparing the baselines to our model on \scs\ are shown in Table \ref{tab:nlu:result}, for both 1-shot and 5-shot setting. Meta Metric-learners achieve the best accuracies over all the methods.  Even with the help of additional sources, the performance of matching network is not better than meta-learner LSTM.

\begin{table*}[h]
\footnotesize
\caption{Comparisons between the models and their baselines on the real-world sentence classification tasks from multiple resources (\scs).}
\label{tab:nlu:result}
\centering
\begin{tabular}{ccccc}
\toprule
\multirow{2}{*}{\textbf{Model}} & \multirow{2}{*}{\textbf{Matching Fn}} & \multirow{2}{*}{\textbf{Additional Data}} & \multicolumn{2}{c}{\textbf{Average Acc}} \\
\cline{4-5}
& &  & 1-shot & 5-shot \\
\midrule
Matching Network & Basic & Y & 53.59\% & 68.73\%   \\
Matching Network & FCE & Y & 54.24\% & 70.28\%  \\
Meta-learner LSTM & - & N & 56.98\% & 72.54\%  \\
Meta Metric-learner & Basic & Y & 57.62\% & 73.83\% \\
Meta Metric-learner & FCE & Y & 58.13\% & 74.54\% \\
\bottomrule
\end{tabular}
\end{table*}

\paragraph{Omniglot}
For Omniglot, each task corresponds to an alphabet, and the total is number of 50. We randomly choose 20 tasks in this experiment. Considering multi-task setting for Omniglot has a clear motivation, as cross-alphabet knowledge sharing is likely to be useful. We use a similar strategy to find related top-$s$ ($s$ is different according to different tasks) tasks and train the metric learner. The following setting is used: splitting each task with 5:2:3 as meta-training, meta-validation and meta-testing. The validation set is used to tune the hyper-parameters.
We use 10 examples per class for evaluation in each test set. Results comparing the baselines to our model on Omniglot are shown in Table \ref{tab:omn:result}. For 1-shot and 5-shot, our model can achieve better classification accuracies than others. With the help of more resources, matching network can beat meta-learner LSTM around 2-3\%.

\begin{table*}[h]
\footnotesize
\caption{Comparisons between the models and their baselines on Omniglot.}
\label{tab:omn:result}
\centering
\begin{tabular}{ccccc}
\toprule
\multirow{2}{*}{\textbf{Model}} & \multirow{2}{*}{\textbf{Matching Fn}} & \multirow{2}{*}{\textbf{Additional Data}} & \multicolumn{2}{c}{\textbf{Average Acc}} \\
\cline{4-5}
& &  & 1-shot & 5-shot \\
\midrule
Matching Network & Basic & Y & 94.62\% & 98.37\%   \\
Matching Network & FCE & Y & 95.84\% & 98.65\%  \\
Meta-learner LSTM & - & N & 93.54\% & 97.22\% \\
Meta Metric-learner & Basic & Y & 95.32\% & 98.56\% \\
Meta Metric-learner & FCE & Y & 95.79\% & 98.83\% \\
\bottomrule
\end{tabular}
\end{table*}

\paragraph{Amazon Reviews}
We treat each product category as a task and the goal is classify each sample into one of four categories. We select one domain as the target few-shot learning task. For the rest tasks, we select 5 tasks from them to train the metric-learner and 1 task used as the meta-validation set to search over hyper-parameters. At each iteration, data from one task is sampled to train the models. Under such condition, Meta-learner LSTM can take the benefit of using additional samples from other tasks. We use 5 sentence per class for evaluation in each test set. The results are shown in Table \ref{tab:sentiment:result}.  The meta-learner attains result that are better than the baselines discussed. For 5-shot, we are able to improved matching network more than 5\%, whereas for 1-shot, the results are still competitive and outperform the second-best around 2\%. It is a little surprising that LSTM meta-learners can not work well on this dataset, with/without additional data, showing that using sample of different sources to jointly train meta-learner is not help.

\begin{table*}[h]
\footnotesize
\caption{Comparisons between the models and their baselines on Amazon Reviews.}
\label{tab:sentiment:result}
\centering
\begin{tabular}{ccccc}
\toprule
\multirow{2}{*}{\textbf{Model}} & \multirow{2}{*}{\textbf{Matching Fn}} & \multirow{2}{*}{\textbf{Additional Data}} & \multicolumn{2}{c}{\textbf{Average Acc}} \\
\cline{4-5}
& &  & 1-shot & 5-shot \\
\midrule
Matching Network & Basic & Y & 44.25\% & 51.92\%   \\
Matching Network & FCE & Y & 47.18\% & 54.64\%  \\
Meta-learner LSTM & - & N & 42.69\% & 51.36\%  \\
Meta-learner LSTM & - & Y & 43.47\% & 52.05\%  \\
Meta Metric-learner & Basic & Y & 48.25\% & 58.44\% \\
Meta Metric-learner & FCE & Y & 49.38\% & 60.82\% \\
\bottomrule
\end{tabular}
\end{table*}

\subsection{Single Task Setting}
To demonstrate the effectiveness of our meta metric learner, we also execute experiments for single task/resource, on Sentence Classification Service and Omniglot datasets, in which no auxiliary set $\mathscr{D}_{aux}$ is available from other tasks. Thus we need to perform $k*2$-shot learning ($k=1,2$), i.e., for each class, we split its samples into two parts equally, to update the model and base learner jointly. Two separate splits are used: 1) 50\%, 20\%, and 30\% classes for training, validation and testing,  2) 30\%, 20\%, and 50\% classes for training, validation and testing. For 1), k$-shot$,$N$-way classification is performed. 2) is a challenging setting since the number of classes in meta-train is smaller than meta-test. We use all classes in $\mathscr{D}_{meta-train}$ and $\mathscr{D}_{meta-test}$, which is different from  $k$-shot,$N$-way setting. For both splits, the validation set is used to adjust the hyper-parameters. To have a fair comparison, all the baselines trained with $k*2$ samples per class according to their own recipes.

For \scs, all 10 tasks are used in the evaluation and the results are shown in Table \ref{tab:nlu:single}. All the results are measured after 10 runs.
It is notable that for 3 vs. 5 split, meta-learner LSTM is not able to be employed.  In both 2 and 4 shot, our model outperforms the other methods.
The performance of meta-learner LSTM is slightly better than Matching Network. The classification accuracy of meta metric-learner with FCE is higher than the basic version, which shows that FCE can improve the basic one on \scs. On the other hand, it is obvious to see that the 3 vs. 5 split is a more challenging task. Comparing the results in both cases, the performance of 3 vs. 5 is around 10\% lower than 5 vs. 3 cases.

\begin{table*}[h]
\footnotesize
\caption{Average classification accuracies on \scs\ of different approaches in single task setting.}
\label{tab:nlu:single}
\centering
\begin{tabular}{cccccc}
\toprule
\multirow{2}{*}{\textbf{Model}} & \multirow{2}{*}{\textbf{Matching Fn}} & \multicolumn{2}{c}{\textbf{5 vs. 3 split}} & \multicolumn{2}{c}{\textbf{3 vs. 5 split}} \\
\cline{3-6}
&  & 2-shot & 4-shot & 2-shot & 4-shot \\
\midrule
Matching Network & Basic & 59.42\% & 68.65\% & 48.15\% & 57.28\%  \\
Matching Network  & FCE & 59.57\% & 68.91\% & 48.74\% & 57.31\% \\
Meta-learner LSTM & Basic & 60.15\% & 69.06\% & - & - \\
Meta Metric-learner & Basic & 60.81\% & 69.14\% & 50.23\% & 58.44\% \\
Meta Metric-learner & FCE & 61.27\% & 69.58\% & 50.56\% & 59.02\% \\
\bottomrule
\end{tabular}
\end{table*}

On Omniglot set, we randomly choose 20 tasks from the whole tasks in the evaluation. Same setting and data split are used as in \scs. The results are reported after 15 runs and described in Table \ref{tab:omn:single}. Similar trend is observed: the performance of meta metric-learners are better than others.
Meta-learner LSTM and matching network achieve almost the same performance for 5 vs. 3 split. Different from \scs, considering the accuracies of two Meta Metric-learner, FCE function seems does not help a lot here.

\begin{table*}[h]
\footnotesize
\caption{Average classification accuracies on Omniglot of different approaches in single task setting.}
\label{tab:omn:single}
\centering
\begin{tabular}{cccccc}
\toprule
\multirow{2}{*}{\textbf{Model}} & \multirow{2}{*}{\textbf{Matching Fn}} & \multicolumn{2}{c}{\textbf{5 vs. 3 split}} & \multicolumn{2}{c}{\textbf{3 vs. 5 split}} \\
\cline{3-6}
& & 2-shot & 4-shot & 2-shot & 4-shot \\
\midrule
Matching Network & Basic & 96.02\% & 96.83\% & 93.94\% & 94.88\%  \\
Matching Network & FCE & 96.50\% & 97.39\% & 94.14\% & 95.26\% \\
Meta-learner LSTM & - & 96.54\% & 97.45\% & - & - \\
Meta Metric-learner & Basic & 97.24\% & 98.33\% & 95.77\% & 96.32\% \\
Meta Metric-learner & FCE & 97.38\% & 98.47\% & 95.69\% & 96.24\% \\
\bottomrule
\end{tabular}
\end{table*}

\section{Conclusion}
In this paper, we proposed a meta metric learner for few-shot learning, which is a combination of an LSTM meta-learner and a base metric classifier. The proposed method takes several advantages such as is able to handle unbalanced classes as well as to generate task-specific metrics. Moreover, as shown in the results, using the meta-learner to guide gradient optimization in matching network seems to be a promising direction. We evaluate our model on several datasets, in both single task and multi-tasks settings. The experiments demonstrate that our approach outperforms is very competitive to the state-of-the-art approaches for few-shot learning.

There are several directions for future work. First, we will focus on selecting the data from related domains/resources to support the training of meta metric learners. Secondly, it would be interesting to propose an end-to-end framework of the meta-learner to leverage the data from different domains/sources/tasks for the training, instead of the current two-stages procedure. Finally, we would like to move forward to apply the current framework in other applications, such as language modeling \cite{DBLP:journals/corr/DauphinFAG16}, machine translation \cite{NIPS2016_6469} and vision applications \cite{DBLP:conf/cvpr/LuKZCJF17}.

\bibliographystyle{unsrt}
\bibliography{main}

\end{document}